%% file: main.tex
\documentclass[11pt]{article}

\usepackage[preprint]{acl}

\usepackage{times}
\usepackage{latexsym}

\usepackage[T1]{fontenc}

\usepackage[utf8]{inputenc}

\usepackage{microtype}

\usepackage{inconsolata}

\usepackage{graphicx}
\usepackage{natbib}
\usepackage{algorithm}
\usepackage{algorithmic}
\usepackage{xcolor}
\usepackage{wrapfig}
\usepackage{tcolorbox}
\usepackage{subcaption}

\usepackage{booktabs}
\usepackage{multirow}
\usepackage{tikz}

\usepackage{bigfoot}
\usepackage[dont-mess-around]{fnpct}
\DeclareNewFootnote{ANote}[fnsymbol]

\makeatletter
\def\thanksANote#1#2{
    \protected@xdef\@thanks{\@thanks
        \protect\footnotetextANote[#1]{#2}}%
}

\title{Emotion-Aware and Context-Grounded Dialogue Agents for Personalized Video Game Interactions}

\author{
 \textbf{Michael Adjeisah\textsuperscript{1,2,\footnotemarkANote}\thanksANote{1}{Work done at SIE.}},
 \textbf{Jin Zhang\textsuperscript{3}},
 \textbf{Pierluigi Vito Amadori\textsuperscript{2}},
 \textbf{Ryan Spick\textsuperscript{2}},
 \\
 \textbf{Eshani Kawshika Fernando\textsuperscript{1}},
 \textbf{Kwame Omono Asamoah\textsuperscript{4}},
 \textbf{Huiwen Zhao \textsuperscript{1}},
\\
 \textbf{Jian Jun Zhang\textsuperscript{1}},
\textbf{Guy Moss\textsuperscript{2,†}\thanksANote{2}{Correspondence: \href{mailto:email@domain}{guy.moss@sony.com}; 
\href{mailto:email@domain}{jchang@bournemouth.ac.uk}}},
 \textbf{Jian Chang\textsuperscript{1,†}}
\\
 \textsuperscript{1}National Centre for Computer Animation, Bournemouth University, UK \\
 \textsuperscript{2} SIE, London, UK,
 \textsuperscript{3}SIE, San Mateo, USA\\
 \textsuperscript{4}Ningbo China Institute for Supply Chain Innovation, MIT Global Scale Network, China 
\\
}

\begin{document}
\maketitle
\input{sections/abstract}
\input{sections/introduction}
\input{sections/related}
\input{sections/dataset}
\input{sections/method}

\input{sections/experiments}
\input{sections/conclusion}

\input{sections/limitation}

\section*{Acknowledgments}
Work done in collaborations SIE. This work was supported by the European Union’s Horizon 2020 research and innovation program under the Marie Skłodowska-Curie grant agreement No. 900025.

\bibliography{custom.bib}

\appendix

\input{sections/appendix}

\end{document}

%% file: sections/abstract.tex
\begin{abstract}
We present \textit{ChatNPC}, a game companion that dynamically personalizes its responses in real time based on players' emotional shifts to enhance user immersion. The model integrates sequential data, leveraging the dynamic nature of streams to benefit from player information, spoken lines, and in-game context. We leverage recent progress in LLMs' tool-calling capabilities to extract vital information from memory and recognize potential constraints for accurate reasoning, thereby tackling the complexity of NPC conversation scenarios. The task is divided into a game sentinel-guided agent (\textit{SeGent}), a memory capability, and a chat planning tool for reasoning instantiation. The approach benefits from a lightweight \textit{game-template} as an information framework, with relevant details and a thorough reasoning layout. We conduct extensive experiments on a newly developed game dataset for in-game context NPC dialogue and demonstrate that ChatNPC sequentially captures players' emotional shifts over time; responses are more naturalistic and human-like with appropriate conversational cues, pauses, and sighs; and utterances remain faithful to the dynamics of in-game context and player actions, supporting narrative continuity.
\end{abstract}

%% file: sections/introduction.tex
\section{Introduction}
\begin{figure}[t]
\centering
  \includegraphics[width=7.5cm]{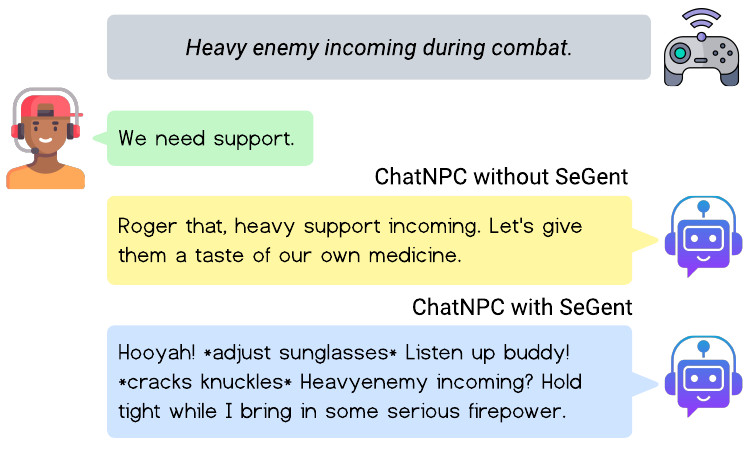}
  \caption{Snippets of ChatNPC responses showing more naturalistic and human-like, with appropriate conversational cues, making it more believable with the \textit{SeGent}.}
  \label{fig:sample_npc_resp}
\end{figure}

Most studies present various applications of LLMs \textit{in} and \textit{for} the broader ecosystem of games, including the different roles they can play within a game \cite{sweetser2024large}. Their theoretical frameworks are extensively based on emulating human behavior to play games at a human or near-human level \cite{liao2024efficacy} and as quest providers for immersive and personalized gaming experiences for game creators that use LLMs as conversational agents \cite{gallotta2024large}.  However, an interactive companion intended to enrich or guide the player experience without competing with or altering the game mechanics remains relatively unexplored. For example, players are allowed to mislead the king into uttering the name of a particular weapon, which eventually materializes in defeating the king in 1001 Nights \cite{sun2023language}. Likewise, in \textit{Gandalf}\footnote{https://gandalf.lakera.ai/}, a player can trick an LLM into exposing a password. A different approach to LLM for user-game immersion could be an interactive agent that does not causally interact with the game world 
\cite{gallotta2024large}. 
ChatNPC is an agent that can interact with the game world without deception or manipulation while attending to subtle player cues 
during gameplay to foster contextual interaction and emotional connection (Figure \ref{fig:sample_npc_resp}).


Non-player characters (NPCs) perform various roles and functions, such as quest-givers, vendors, shopkeepers, allies and companions, enemies and adversaries, and supporting characters \cite{gallotta2024large, weir2022ontologically, uludaugli2023non}. They hold critical information about the game to enrich the player’s experience by adding to the atmosphere of the game world and making it more believable \cite{croissant2023emotion, uludaugli2023non}.
Developing a virtual agent that engages in meaningful and contextually appropriate dialogue with players throughout gameplay, emphasizing the importance of stealth and situational awareness, can provide essential information, plot developments, and mythology that enrich the game world and drive player immersion. 

In this work, we present an adaptive game companion that dynamically personalizes its responses based on player utterances, in-game context, and real-time emotional shifts, thereby fostering naturalistic and emotive interactions that enhance immersion and engagement within the broader gameplay. This includes emotional responses during gameplay, such as those triggered by exploration, combat, or decision-making moments. For example, ChatNPC can react differently depending on whether the player is excited after achieving a milestone or frustrated during a difficult challenge. 

The model integrates sequential data from audio streams (spoken lines), player information (health, level), emotions expressed in the spoken lines, in-game context, and an inner monologue. Specifically, the cues in the audio streams are extracted and interpreted by an agent: \textit{SeGent} (Section \ref{sec:method}). It integrate an encoding mechanism into the model architecture to analyze and interpret player emotional shifts and understand the situational aspects of the gameplay by tracking previous context and current context information to initiate on-the-fly prompt adjustments
to trigger the LLM (a \textit{chat planning agent}) inference to generate responses (Figure \ref{fig:sample_npc_resp}). 
We design a memory mechanism \cite{packer2023memgpt} to track events occurring during the playthrough and interactions with the player and leverage recent progress in LLM tool calling \cite{kim2023llm} capabilities to extract vital information from memory during problem distillation \cite{yang2024buffer} 
for accurate reasoning \cite{yang2024buffer, wei2022chain}.
To foster a dynamic and diverse conversation, we guide the chat planning agent using a well-defined agent \textit{persona} \cite{jiang2024evaluating} that includes knowledge and communication style. 
Our contributions are as follows:
\begin{enumerate}
    \item We present \textit{ChatNPC}, a game companion capable of personalizing its responses in real-time based on players' emotional shifts. 
    
    \item We employ a novel application of a sentinel-guided agent to foster emotional connection and manage the game's contextual elements for a seamless narrative flow.

    \item We design a \textit{memory agent} to process information across the \textit{in-context} and \textit{out-of-context} windows, keeping track of events during the playthrough for reasoning, instantiation, and rescaling of thought.

    \item  We conduct extensive experiments on newly developed game datasets for in-game context NPC dialogue.
\end{enumerate}

%% file: sections/related.tex
\section{Related Work}\label{sec:related}
LLMs are inherently suited for natural language dialogue (NLD); therefore, they are typically presented as conversational agents, displaying remarkable skills in memory \cite{packer2023memgpt,sumers2023cognitive}, tool usage \cite{yang2024gpt4tools,schick2024toolformer,ruan2023tptu,qin2023toolllm} and planning \cite{yang2024buffer, wei2022chain}, often leading researchers to give them reasoning and creativity qualities \cite{gallotta2024large, wei2022chain, zhao2024expel,zhang2024generative}. The growing capabilities have motivated recent efforts to incorporate them as game agents \cite{Oliver_Mateas_2021} with believable proxies of human behavior to empower interactivity in immersive environments.

\begin{table*}
  \centering
   \small
  \begin{tabular}{ll}
    \hline
    \textbf{Spoken-line:}    &\textcolor{cyan}{We need support}  \\
    \textbf{In-game Context:}  &\colorbox{lightgray}{Heavy enemy incoming during combat} \\
    \textbf{Response:} &\textcolor{blue}{Roger that, heavy support incoming. Let's show them what we've got!} \\
    \textbf{Spoken-line:}    &\textcolor{cyan}{That generator might be the distraction I need.}  \\
    \textbf{In-game Context:}  &\colorbox{lightgray}{Hangar Mission - Escape hangar interaction prompt} \\
    \textbf{Response:} &\textcolor{blue}{Oh man, you are right! That generator could be exactly what we need to get out of here alive.} \\
    \hline
  \end{tabular}
  \caption{\label{dataset-sample}
    Sample curated dataset from various games with desired responses for training and testing.
  }
\end{table*}

\subsection{LLMs in Game}
LLMs operate within games as a player \cite{toshniwal2022chess,li2022emergent,meta2022human,yao2020keep}, NPCs \cite{gallotta2024large, weir2022ontologically, li2022emergent}, an assistant providing hints \cite{akata2023playing,xu2023exploring},  a game master \cite{zhu2023calypso} controlling the flow of the game and acting as a commentator \cite{ranella2023towards} of an ongoing play session. Park et al. \citeyearpar{park2023generative} introduced generative agents that simulate believable human behavior, such as daily activities (waking up, cooking breakfast, and going to work). The authors used LLMs to populate an interactive virtual Smallville sandbox environment with 25 generative agents by storing an entire history of the agents’ experiences, synthesizing those memories over time into higher-level representations, and dynamically retrieving them for behavior planning. Other works include using vision LMs (VLMs) to analyze video frames and predict the next steps in the Mario video game \cite{lin2023mm} and to evaluate their effectiveness in the game StarCraft II \cite{ma2023large}. To understand how LLMs behave in interactive social settings, Akata et al. \citeyearpar{akata2023playing} proposed using behavioral game theory to study agents' cooperation and coordination behavior by repeatedly engaging different language models as agents in games with each other in human-like scenarios. What is most common among these techniques is the ability for users to interact with these agents in an NLD.

\subsection{LLMs and NPC Dialogue}
An LLM-powered framework for quests and dialogue generation that places the player at the core of the generative process has been explored, incorporating context awareness \cite{ashby2023personalized,muller2023chatter} and personality modeling \cite{muller2023chatter,latouche2023generating} to create fluent, unique, and engaging dialogue. ChatGPT has recently been used to generate game dialogue by allowing it to take control of NPCs and interact dynamically with players \cite{zhou2023dialogue}. In addition to generating NPC dialogue, many research works integrated LLMs into game mechanics using quest datasets \cite{ashby2023personalized,weir2022ontologically,vartinen2022generating,van2021fine,gao2023turing}. A Minecraft player interacting with a Codex-powered NPC in question-answering and task-completion scenarios demonstrates that conversational prompts can power a conversational agent to generate natural language \cite{volum2022craft}. These approaches greatly benefit both the gaming industry and its global community, as they unleash the immersive and enjoyable experience for users \cite{akoury2023framework}. Regardless, the role of NPCs in games extends far beyond serving as quest-givers, answering questions, and completing tasks. In addition, modern gaming goes beyond just winning and voice commands; players seek a sense of immersion and connection with the virtual world.

\begin{table}
  \centering
  \small
  \begin{tabular}{llll}
    \hline
                             & \textbf{Utterances}   & \textbf{Context}  & \textbf{Responses}  \\
    \hline
    \hline
    Token counts   &  148668   &   156616   &   497319   \\
    Unique words   &   7815    &   1758   &    8248 \\
    \hline
    \hline
    \multicolumn{4}{c}{\textbf{Train} \hspace{1cm} \textbf{Validation} \hspace{1cm} \textbf{Testing}}  \\
    \multicolumn{4}{c}{\hspace{-0.5cm} 10,000 \hspace{1.3cm} 3,885 \hspace{1.5cm} 512}  \\
    \hline
  \end{tabular}
  \caption{Statistics of the iGCD Dataset.} 
  \label{data-statistics}
\end{table}

\subsection{Player Emotions in Games}
Many studies explore the emotional responses evoked during gameplay and interactions with NPCs \cite{marincioni2024effect}. A survey by Mozikov et al. \citeyearpar{mozikov2024good} focused on the decision-making of LLMs and their alignment with human behavior in emotional states within various strategic games. They revealed that emotional prompting, particularly for certain emotions (such as anger), can disrupt some LLMs' "superhuman" alignment, similar to how human emotional responses do. Language models have been used to extract emotion scores from players' emotional reactions to the game by capturing interactions with the NPC through user input, thereby understanding the emotional dynamics within gaming environments \cite{marincioni2024effect}. 
Current research has yet to fully explore the potential of an LLM-powered game companion that apprehends players' emotional shifts during gameplay. Hence, we present ChatNPC, a naturalistic and emotive game-powered dialogue agent.

\begin{figure*}[t]
\centering
  \includegraphics[width=12cm]{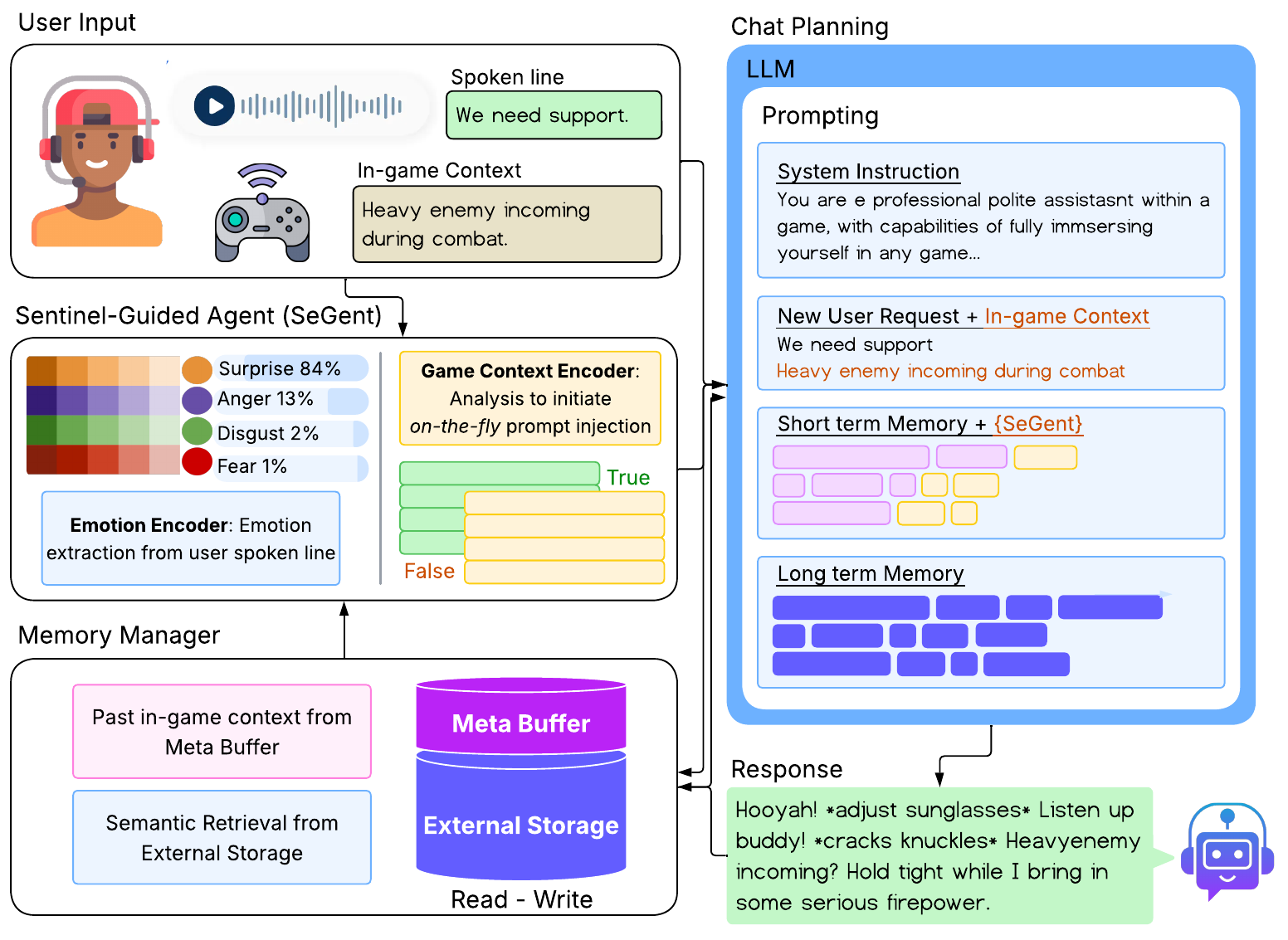}\\
  \caption {The overall architecture of ChatNPC consists of three main modules. The sentinel-guided agent with two different encoding types (an \textit{in-game context} and \textit{emotion} encoders), the memory manager to process information between the \textit{in-context} and the \textit{out-of-context} window, and the \textit{chat planning agent} to format the output request.}
    \label{fig:method}
\end{figure*}

%% file: sections/dataset.tex
\section{Dataset Curation}
In this work, we introduce a new \textbf{i}n-\textbf{g}ame \textbf{c}ontext \textbf{d}ialogue (iGCD) dataset to evaluate the effectiveness of the game companion.
The dataset consists of scripted speech (spoken lines) and in-game mechanics from four game types: adventure, action-adventure, first-person shooter, and vehicular combat, provided by a commercial game studio from their production archives. They are mainly dialogue written by game developers as part of the game's narrative and associated with in-game mechanics. This makes the utterances the original character dialogue from the games. 
We process the data by filtering for duplicates and one-word utterances, except for specific (inappropriate) words. For instance, we remove duplicate instances if a speech appears more than once within the same context.
We, however, retain it if the same speech appears in different contexts, since its meaning 
differs depending on the game scenario.

As the dataset does not come with a desired output, we initially generate responses through instruction fine-tuning Llama 2 70B using the Open-Platypus \cite{lee2023platypus} 
dataset\footnote{https://huggingface.co/datasets/garage-bAInd/Open-Platypus} to produce contextually appropriate results. We use few-shot prompting with examples from game dialogue to guide style at this stage. These preliminary responses underwent a rigorous evaluation process involving AI-based assessment (LLM-as-Judge) 
and 
human-in-the-loop review 
to ensure quality.
The human evaluators were tasked with reviewing 20\% (2,512 examples) due to the large number of pairs. These were randomly sampled and stratified by genre. Within this, 15\% were added back to the training set after human editing, and 5\% (512 examples) were held out as 'gold responses' for testing. A response is termed a "\textit{gold response}" when, after human editing, it meets all four quality criteria (naturalness, relevance, coherence, and expressiveness) to a human-quality standard (Appendix \ref{sec:appendix} ). 
All names are preserved in the dialogue dataset with a special token $<\textit{player}>$. The overall dataset comprises 14,397 pairs of utterances, in-game context, and corresponding responses, with 10,000 used for training, 3,885 for validation, and 512 for testing.  Samples of the collected dataset are given in Table \ref{dataset-sample}, and statistics are shown in Table \ref{data-statistics}. 

%% file: sections/method.tex
\section{Method} \label{sec:method}

\subsection{Sentinel-Guided Agent}\label{subsec:SenEmb}
As shown in Figure  \ref{fig:method}, the \textit{SeGent} seamlessly integrates two distinct sentinel-guided encoder types: an \textit{in-game context} and an \textit{emotion} encoder.\\


\noindent
\textbf{In-game Context Encoder} focuses on understanding the situational aspects of gameplay by considering the game's storyline and environmental factors to ensure cohesion. It tracks previous and current context information to effectively initiate \textit{on-the-fly} prompt adjustments to model responses based on a "true/false" validation. To achieve this, we use an embedding model $\phi(\cdot)$ to capture the difference between the current and past context in the game and compute the cosine similarity. For each input $i$, we compare the current in-game context $C_i$ by computing the embedding similarity between the previous in-game context $C_{i-1}$ and $C_i$ as:
\begin{equation}
    \mathrm{\mathbf{pr}}_{adj} =
  \mathbf{1}_{[\mathcal{T},1]}\Big(\mathrm{Sim}\big(\phi(C_i),\phi(C_{i-1})\big)\Big)
\end{equation}
where $\mathcal{T}$ is the threshold (0.8 $\sim$ 0.9 is recommended) to determine whether the in-game context matches for on-the-fly prompt adjustment $\mathrm{\mathbf{pr}}_{adj}$, and $\mathbf{1}_{[\mathcal{T},1]}$ denotes the characteristic function of the interval $[\mathcal{T},1]$.
The injected prompt provides clear instructions within the system prompt to guide the LLM in reacting when new user input is received. 
\\

\noindent
The \textbf{Emotion Encoder} analyzes and interprets users' emotions by processing spoken lines and subtle cues. It extracts emotional signals as metadata, which can inform the LLM's reasoning. The encoder is based on a zero-shot classification model pre-trained on natural language inference (NLI) and further trained on the GoEmotions dataset, allowing it to identify 28 distinct emotional categories \cite{demszky2020goemotions}.
Overall, the spoken line $\mathcal{S}$ is converted into an embedding vector $V_{\mathcal{S}}$ for a semantic representation using the same embedding space as the emotion vectors $E_\mathcal{J}$ as: 
\begin{equation} \label{eq2}
    \mathrm{meta}_{emo} = \mathrm{argmax}_\mathcal{J}\Big(\mathrm{Sim}(V_{\mathcal{S}}, E_\mathcal{J}) \Big) ,
\end{equation} 
where $\mathcal{J}\in\{j_1, j_2, ..., j_n \}$ is the vector for emotion labeled in the dataset. The emotion vector with the highest similarity score determines the most likely emotion expressed.

\subsection{Memory Manager} \label{subsec:memAgent}
The \textit{memory manager} module processes information between the \textit{in-context} and \textit{out-of-context} windows to keep track of events \cite{packer2023memgpt}. The \textbf{in-context window}
consists of the system instructions and a \textit{meta-buffer} for storing a rolling chat history.
The system instruction is 
a thoroughly drafted piece of information for reasoning instantiation. In this work, the game-specific system prompt includes role specification, context definition, and task description (\textit{game-template}), input details, output requirements, constraints and rules, and examples to enhance clarity. The \textit{meta-buffer} stores the conversation between ChatNPC and the player, using the FIFO (First In, First Out) operation to append and remove interactions dynamically. 
We keep track of the current number of tokens in the in-context window. When the system receives a new input, the incoming messages are appended to the \textit{meta-buffer}. It then concatenates to the primary prompt tokens to trigger the LLM inference for the response. The oldest interactions in the queue are then moved to \textit{archival storage} when the primary prompt exceeds predefined limits.

\textbf{Out-of-context window} utilizes Chroma Database (DB) to store information beyond the primary context window, enabling efficient retrieval and continuity over a long horizon of conversational interactions. We use paging to store and update memories efficiently by setting an arbitrary threshold to control the maximum number of long-term memories stored \cite{packer2023memgpt}. Finally, we implement a semantic matching \cite{rao2019bridging} to identify the core meaning behind the text, rather than exact word matches, to retrieve out-of-context data. We also retrieve relevant information as a pre-thought mechanism to improve ChatNPC-player dialogue. The pre-thought process uses historical data to determine whether the task has already been completed or previously initiated. This approach proves especially practical when the player has previously interacted with or completed certain levels within the game (see Ablation: Impact of Memory Agent \ref{subsec:ablation}). A novel integration of these techniques pushes toward a production-ready pipeline for game dialogue.

\begin{table}
  \centering
  \begin{tabular}{lll}
    \hline
                                         & \multicolumn{2}{c}{\textbf{Context Window}}\\
    \textbf{Model (Open-source)}           & \textbf{Token} & *\textbf{Messages}\\
    \hline
    \hline
    Llama 2                 & 4k      &   50         \\
    Llama 3                 & 8k      &   50         \\
    \hline
    Mythalion 13B           &  4K    &   50         \\
    Roleplay Llama 3 8B     &  8K     &   50         \\
    \hline
  \end{tabular}
  \caption{Llama 2\&3 open-source models, two different roleplay models based on Llama 2\&3, and the primary context window length. The default maximum token of *messages is set to an average of 50 tokens (approximately 250 characters).} 
  \label{context-window-guide}
\end{table}

\subsection{Chat Planning Agent} \label{subsec:chatPA}
The \textit{chat planning} agent is responsible for formatting the output.  When a new user query is fired, information from memory: \textit{meta-buffer} and \textit{archival storage} (if necessary), and \textit{SeGent} information will be passed to the problem distiller to extract critical state and contextual information along with relevant constraints for reasoning instantiation.  The problem distiller, similar to the buffer of thoughts (BoT) \cite{yang2024buffer}, focuses on extracting key elements from the input task. 
A \textbf{Game-template} tailored for the companion is designed as an information framework that enriches the agent's ability to support and enhance the player's experience in the game world. The template includes details about the game's storyline, objectives, and 'tutorial and hints', allowing the agent to remain aware and provide timely, contextually appropriate feedback to the player. \\
\begin{table*}
  \centering
  \small
  \begin{tabular}{llllllll} 
    \hline
      &              & \multicolumn{3}{c}{\textbf{ROUGE-L}} & \multicolumn{3}{c}{\textbf{ROUGE-WE}}\\
      &\textbf{Model}  & \textbf{Precision} & \textbf{Recall}& \textbf{F1} & \textbf{Precision} & \textbf{Recall}& \textbf{F1}\\
    \hline
    \hline
    \multirow{5}{3em}{Base} &Llama 2 7B &  0.327 & 0.25 & 0.275 & 0.465 & 0.417 & 0.443 \\
    &Llama 2 13B   & 0.383 & 0.272 & 0.318 & 0.48 & 0.436 & 0.461       \\
    &Llama 2 70B   & 0.516&0.404&0.453&0.599&0.55&0.573 \\
    \cmidrule(l){3-8}
    &Llama 3 8B      &0.391 & 0.279 & 0.326 & 0.489 & 0.441 & 0.482  \\
    &Llama 3 70B    & 0.523&0.419&0.468&0.621&0.571&0.595 \\
    \hline
    \hline
    \multirow{2}{4em}{Roleplay}& Llama 3 8B &  0.553 & 0.481 & 0.51 & 0.682 & 0.607 & 0.648 \\
    &Mythalion 13B       &  0.51 & 0.427 & 0.449 & 0.629 & 0.58 & 0.602 \\
    \hline
    \hline
    \multirow{2}{5em}{Fine-tuned} & Llama 2 70B      & \textbf{0.783} & \textbf{0.64} &\textbf{0.687}&\textbf{0.855}&\textbf{0.805}&\textbf{0.83} \\
    &Llama 3 70B      & \textbf{0.807} & \textbf{0.661} & \textbf{0.715} & \textbf{0.883} & \textbf{0.836} & \textbf{0.862} \\
    \hline
  \end{tabular}
  \caption{\label{main-results}
    The overall model performance is based on Llama base, roleplay, and fine-tuned models. ChatNPC responses are scored against the human-in-the-loop 'gold' responses. 
  }
\end{table*}
\textbf{Agent persona}. To maintain a successful, diverse conversation, we prompt the \textit{chat planning} agent with a character role, a \textit{persona} that includes personality, knowledge, and communication style. We use the character's iconic style of expression/talking in the game world as a pre-conversation. Each time a dialogue request is made, an arbitrary line from the pre-conversation is concatenated with the prompt tokens to generate a faithful response from the companion.   
Additionally, ChatNPC relies on an inner monologue
to create a more conscious AI that guides its reactions by attempting to reason the best way to respond. 
By combining these functionalities, ChatNPC performs multi-step operations and computations, enabling it to handle complex conversational scenarios and provide more accurate, detailed responses. The detailed prompt for ChatNPC is included in Appendix \ref{sec:prompt}.

%% file: sections/experiments.tex
\section{Experiments} \label{sec:exp}
\subsection{Experimental Settings}

\paragraph{LLM}
We use open-source models, including the Llama baseline models as shown in Table \ref{context-window-guide}, and two role-play models: Roleplay Llama-3-8B\footnote{https://huggingface.co/vicgalle/Roleplay-Llama-3-8B},  
and Mythalion 13B\footnote{https://huggingface.co/PygmalionAI/mythalion-13b} model based on Llama 2 13B. For supervised fine-tuning, we utilize both Llama 2\&3 70B. We dedicate the Llama 2 as the backbone of our ChatNPC, including both the main experiment and the ablation study (see Subsection \ref{subsec:ablation}).

\subsection{Implementation Details}
\paragraph{LLM Instruction Fine-tuning}
Given that our dataset does not contain a desired output, we initially generate responses via instruction fine-tuning Llama 2 70B, engaging Low-Rank Adaptation (LoRA) \cite{hu2021lora} using the Open-Platypus \cite{lee2023platypus} dataset to produce contextually appropriate results. We set LoRA rank to 16, alpha to 32, and dropout to 0.1 for 15,000 training steps, with a batch size of 16, a learning rate of 4e-5, and a weight decay of 0.01. This configuration updates only 0.1\% of model parameters, preserving base knowledge while adapting to game dialogue. We trained the model to generate structured JSON outputs conforming to a Pydantic schema, ensuring machine-readable responses. The results underwent a rigorous evaluation process, including automated AI-based assessment \cite{dalal2024inference} and human-in-the-loop evaluation \cite{elangovan2024considers}. 
We finally instantiate similar parameters for \textit{supervised fine-tuning} on iGCD with the desired output to demonstrate how the underlying dataset improves ChatNPC performance.

\paragraph{Inference and Evaluation}
During generation, we use a base temperature of 0.5 with a random variation of ±0.05, top-k=9, top-p=0.6, and beam search (beams=3) to balance coherence and variety. Maximum response length is capped at 50 tokens with a repetition penalty of 1.2 to prevent loops. For inference with the base LLMs and the roleplay models, we employ a few-shot \cite{brown2020language} prompting approach to facilitate in-context learning. 
We benchmark ChatNPC against the baselines (see Table \ref{main-results}) with the "gold response." 
We use metrics such as \textit{ROUGE-L} and \textit{ROUGE-WE} scores to reference the highest achievable score when comparing the generated results with the gold responses. 
They serve as a development metric for response relevance.
Hyperparameter settings, metrics, and inference details are added in Appendix \ref{sec:appendix} and \ref{sec:appendix-B}.

\subsection{Results and Analysis}
As shown in Table \ref{main-results}, ChatNPC achieves improved overall performance through supervised fine-tuning compared to baselines that use few-shot prompting and role-play models. Specifically, fine-tuned ChatNPC surpasses the base and role-play models by 26.5\% and 20.1\% on ROUGE-WE scores, respectively, with a significant increase in dialogue diversity. Hence, a better interpretation of how ChatNPC effectively addresses the complexity of NPC conversation scenarios. This improvement is attributed to contextual information from the training dataset and the modules instantiated. 
The base LLMs perform poorly mainly because they tend to ignore (not always) the conversational level of interaction. 
Even with few-shot prompting and role-play, models cannot learn the contextual information in task-specific cases.

The increase in scores indicates that curating a high-quality dataset is recommended instead of few-shot learning in task-specific or performance-sensitive applications \cite{hu2021lora}. 
Also, from table \ref{main-results}, we observe no significant differences in the base Llama 2 and 3 70B. We attribute this to the evaluation metrics, as the ROUGE-WE capture matches based on meaning rather than on exact word matches. On the other hand, the curated dataset plays a significant role in ensuring that the model is fine-tuned on highly relevant, diverse, and in-game context-specific instances, thereby facilitating better generalization. Player experience is supported by human evaluation in subsection \ref{subsec:user_exp}. 

The \textit{SeGent} employs a low-computation mechanism, therefore imposing no additional computational overhead on the model architecture. The In-game Context Encoder uses a helper function to generate an embedding for a given text using a pre-trained Sentence Transformer model\footnote{https://huggingface.co/nreimers/MiniLM-L6-H384-uncased} and compute the cosine similarity between the current and previous context. The emotion encoder uses a mechanism distilled from a zero-shot emotion classification model\footnote{joeddav/distilbert-base-uncased-go-emotions-student}. 
In general, the \textit{SeGent} maintains continuity and helps avoid narrative breaks; the memory agent retrieves relevant memories for self-improvement. See subsection \ref{sec:infer_analysis} for inference time analysis.

\subsection{Ablation Study}\label{subsec:ablation}

\begin{figure}[t]
\centering
  \includegraphics[width=5.5cm]{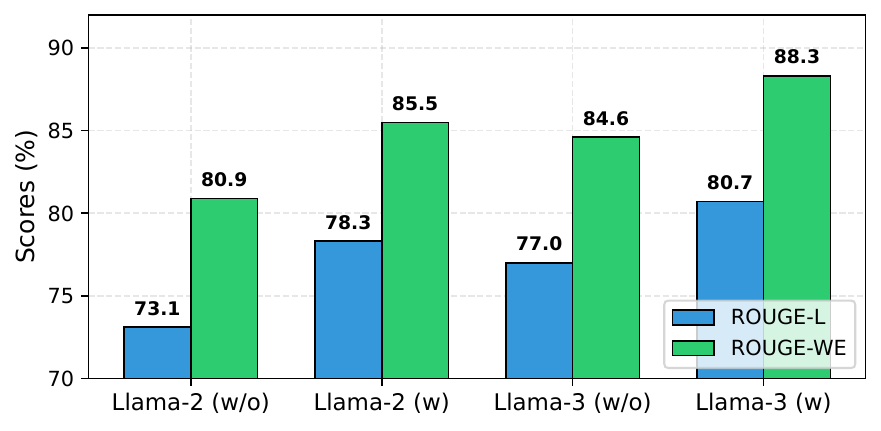}
  \caption{Ablation study on the in-game context (\textit{with} (w) and \textit{without} (w/o)) on both Llama 2 and 3 70B models.}
  \label{fig:in-game_abla}
\end{figure}

\paragraph{Impact of In-game Context Encoder}
Figure \ref{fig:in-game_abla} shows the impact of the in-game context sentinel.
We compare several ChatNPC scenarios \textit{with} and \textit{without} the context sentinel. For this experiment, we sampled from the gold response using the same context. Although ChatNPC without a context sentinel provides contextually appropriate responses, there is a slight deviation in maintaining narrative continuity. 
Also, we see a noticeable decline in performance when the in-game context sentinel is disabled for our fine-tuned Llama 2 and 3. When enabled, responses are more verbose. In a dynamic context change (often in games), 
these adjustments help adapt to the new situation and avoid contradictions or narrative breaks.\\

\begin{figure}[t]
\centering
  \includegraphics[width=7cm]{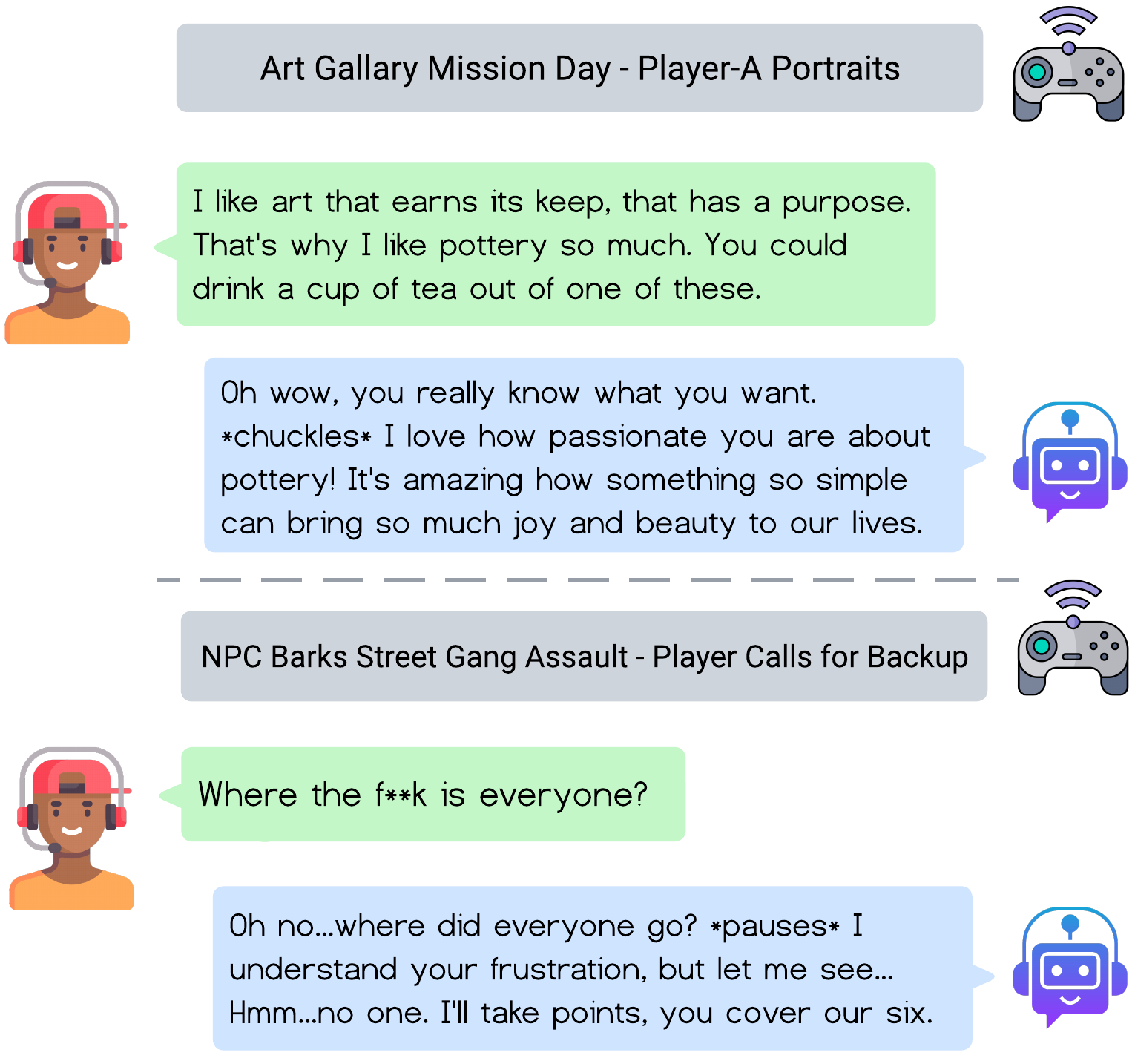}
  \caption{Snippets of the ChatNPC responses when noticing that the user is distressed and moments of sensation.}
  \label{fig:sample_npc_emo}
\end{figure}

\paragraph{Impact of Emotion Encoder.}
The player-agent interaction in Figure \ref{fig:sample_npc_emo} demonstrates that the extracted emotional metadata enables the model to adjust its reactions with precision to the player's emotional state, thereby achieving better alignment and reflecting an understanding of the user's thoughts and feelings over time. This emotional modulation helps create a more personalized, compassionate background by tailoring responses to moments of frustration, excitement, or contemplation. As illustrated, the interaction shows ChatNPC offering words of encouragement when the player expresses frustration and celebrating moments of sensation. This module ultimately helps establish a more human-like connection with the player, fostering a responsive environment.\\

\begin{figure}[t]
\centering
  \includegraphics[width=6cm]{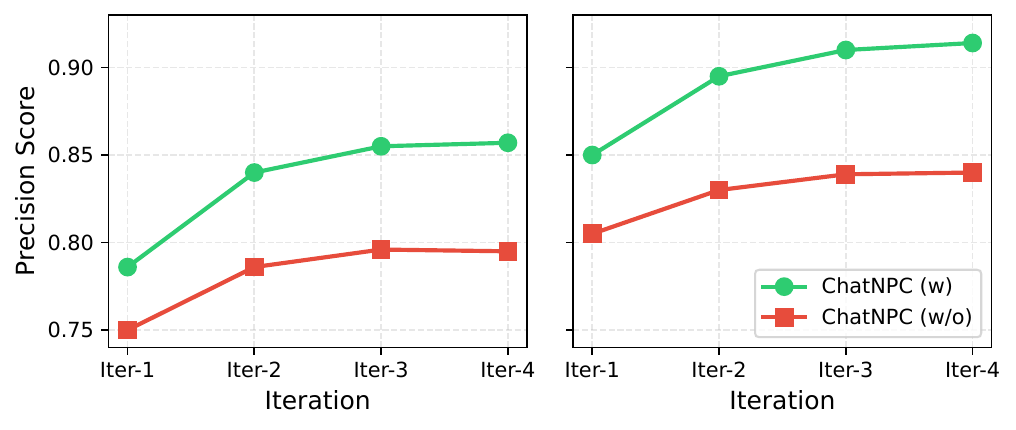}
  \caption{Ablation study on the rescale mechanism by iterating the model (Llama 2 70B as the backbone) multiple times to signify previously interacted situations within the game. ROUGE-L (\textbf{left}) and ROUGE-WE (\textbf{right}).}
  \label{fig:rescale}
\end{figure}

\paragraph{Impact of Memory Agent}
ChatNPC not only leverages memory for relevant information but also rescales (pre-think and improve mechanism) it to perform well by intuition without conscious reasoning \cite{li2023mot}. To understand the potential improvement in the ChatNPC conversation, we sampled 10\% of the test set and iterated ChatNPC (with/without rescaling) multiple times to illustrate previously interacted scenes. At each iteration, we prompt the model to use memory for pre-thinking and improvement. As shown in Figure \ref{fig:rescale}, ChatNPC (-with) consistently improves the ROUGE scores in every iteration. 


\subsection{User Experience}
\label{subsec:user_exp}

\begin{figure}[t]
\centering
  \includegraphics[width=4.5cm]{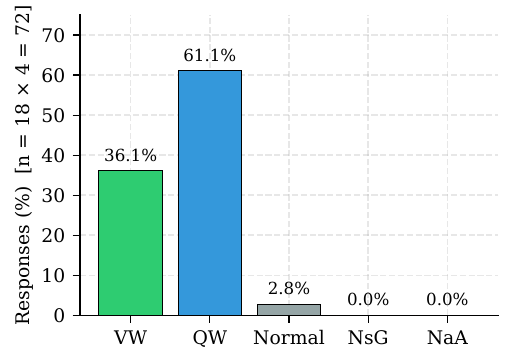}
  \caption{
  Distribution of all responses pooled across Q3–Q6 (n = 18 participants × 4 questions = 72 responses)}
  \label{fig:average_score}
\end{figure}
To understand the interactivity and impressiveness of ChatNPC, we conducted a user experience study involving 18 participants, each engaging in a session with ChatNPC. Following the session, participants are tasked to complete a short questionnaire designed to capture their subjective experience. 
Distribution of all responses pooled across Q3–Q6 (n = 18 participants × 4 questions = 72 responses). VW = Very well; QW = Quite well; NsG = Not so good; NaA = Not at all. All 18 participants rated the experience positively across emotional cues, coherence, human-likeness, and personal relevance, as shown in Figure \ref{fig:average_score}. Individual question results are included as part of Appendix \ref{sec:Questionnaire}. 

\begin{table}
  \centering
  \small
  \begin{tabular}{llllll} 
    \hline    
                    & \multicolumn{3}{c}{\textbf{ROUGE-WE}} \\
      \textbf{ChatNPC}  & \textbf{Precision} & \textbf{Recall} & \textbf{F1} & \textbf{Time (Sec)}\\
    \hline
    \hline
    LLM agent  & 0.857 & 0.8051 & 0.839 & 14-20\\
    Ours & \color{blue} 0.855 & \color{blue}0.805 & \color{blue}0.83 & \color{blue}\textbf{5-8}\\
    \hline
  \end{tabular}
  \caption{\label{compute}
    Computational analysis of ChatNPC average response time with an LLM agent as an emotion summarizer and our \textit{emotion} encoder.}
\end{table}

\subsection{Inference Time Analysis} \label{sec:infer_analysis}
To compare the response time of ChatNPC during inference, we engage an LLM agent as an emotion summarizer, which involves analyzing and condensing emotional cues from the spoken lines. While effective, it requires additional processing steps. Our \textbf{emotion encoder} uses a mechanism distilled from a zero-shot classification model, which imposes no additional computational overhead. Table \ref{compute} indicates that the emotion summarizer does not lead to statistically significant differences in system responses. 
The average response time is 2x that of our emotion encoder. 

%% file: sections/conclusion.tex
\section{Conclusion}
In this work, we propose \textit{ChatNPC}, a game companion that observes and recognizes players' emotional shifts and adjusts its responses accordingly in complex NPC conversation scenarios. Specifically, ChatNPC comprises three modules: a novel game sentinel-guided agent (SeGent) to effectively manage the emotional and contextual aspects of NPC-player dialogue, a memory agent to keep track of events in the playthrough for reasoning instantiation, and a chat planning agent that benefits from game-template and character persona for formatting the output request. We introduce a newly curated iGCD dataset for experimental analysis. The results indicate that ChatNPC sequentially captures players' emotional shifts over time; responses are more naturalistic and human-like, with appropriate conversational cues, pauses, and sighs; and utterances remain faithful to the in-game context and supporting narrative continuity.

%% file: sections/limitation.tex
\section*{Limitations}
\label{sec:ChatNPC-limit}
In gaming scenarios, dynamic emotional shifts over time complicate the accurate expression and recognition of emotions. Addressing this complexity requires a nuanced analytical approach. Currently, ChatNPC relies exclusively on extracting emotions from user speech as metadata. However, spoken lines alone may not sufficiently capture the full range of emotional changes. Integrating facial information with user utterances can yield richer contextual metadata, thereby improving the understanding of a player's emotional state during gameplay. The present work models player emotions as discrete, single-state representations. In contrast, employing a dimensional spectrum could better capture transitional subtleties and facilitate more adaptive, context-aware interactions. Additional information from the in-game environment, such as environmental elements, can further contextualize large language model (LLM) inference, enhancing the interpretation of game scenes from images and video. For example, when a player requests cover, ChatNPC responds with "\textit{We need to take cover behind those crates}," which underscores the importance of detecting scene objects to provide details such as the precise location of the crates. Integrating these data sources with in-game context can substantially improve the agent's responsiveness.

It is also observed that the dataset exhibits an imbalance across game genres, with Vehicular Combat and Action genres accounting for the majority of dialogue pairs (Table~\ref{data-portion}). Future research will address this limitation by collecting additional dialogue from under-represented genres and by exploring rebalancing strategies, such as targeted data augmentation or genre-aware sampling.

Although 18 participants make up a limited sample to generalize our findings on human experience, the current evaluation is an initial pilot study rather than a definitive claim on improvement in user experience. The current results, while preliminary, offer promising signals that justify this expanded investigation.  Also, the ChatNPC response latency is 5–8 seconds, exceeding the acceptable threshold for real-time gameplay interaction. Llama offers strong generative capabilities, but its inference speed limits its use in time-sensitive applications. Future work will therefore consider frontier models optimized for low-latency deployment. It is also important to state that dialogues are one-on-one conversations between the player and a single NPC. Multi-party NPC interactions are left for future work.

%% file: sections/appendix.tex
\section{Appendix A}
\label{sec:appendix}

\subsection{Dataset and Evaluation}
\label{sec:appendix-A1}
The dataset primarily features original game-character dialogue created by developers as part of the narrative, along with associated in-game mechanics. These were extracted from four game genres: adventure, action-adventure, first-person shooter, and vehicular combat.
Inputs are based on real game data, ensuring the model learns to respond to authentic scenarios and mechanics. Targeted human evaluation and editing reduce 'modelese' (synthetic stylistic artefacts) and increase the naturalness, relevance, coherence, and expressiveness of responses. The wide range of genres mitigates the risk of overfitting to a single conversational style. Table \ref{data-portion} reports the proportion of dialogue data per genre. The train, validation, and test splits were stratified by genre to ensure each split maintained the dataset's original genre distribution.
\begin{table}[H]
  \centering
   \small
  \begin{tabular}{ll}
    \hline
    \textbf{Genre}  & \textbf{Count}  \\
    \hline
    \hline
    Vehicular Combat	& 7864 \\
	Action	            & 5840 \\
	Action-Adventure	& 406 \\
	FPS	                & 287 \\
    \hline
  \end{tabular}
  \caption{\label{data-portion}
    Game data distribution by genre.
  }
\end{table}

\begin{figure}[t]
  \includegraphics[width=\columnwidth]{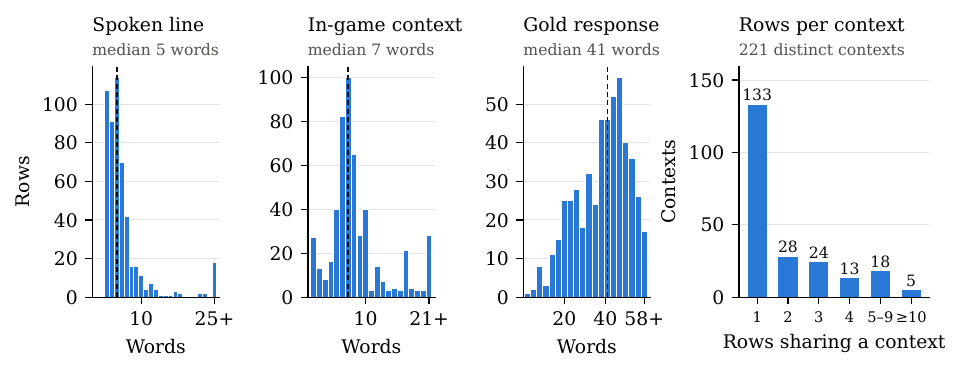}
  \caption{Characteristics of the iGCD test set (512 rows). Left to right: words per player utterance, per in-game context, and per gold response (dashed line: median; the last bar gathers all longer rows), and the number of contexts by how many test rows share them.}
  \label{fig:data_charac}
\end{figure}

Figure \ref{fig:data_charac} summarizes the held-out test set, which consists of 512 utterance–context pairs with gold responses. Player utterances are brief; the median spoken line contains 5 words (interquartile range 4–7), and 91\% contain ten words or fewer, reflecting the concise, reactive speech characteristic of live gameplay. In-game contexts are also compact, with a median length of 7 words (interquartile range 6–10). Gold responses are substantially longer, with a median of 41 words (interquartile range 29–48), approximately eight times the length of the corresponding utterance. Consequently, the companion must generate a comprehensive, contextually appropriate reply from minimal spoken input, making the in-game context and emotion signal the primary sources of information for response generation.

The test set includes 221 distinct in-game contexts. The distribution is long-tailed: 133 contexts (60\%) appear only once, 65 appear two to four times, 18 appear five to nine times, and 5 appear ten times or more (with a maximum of 22 occurrences). Although most contexts are unique, 74\% of the test rows (379 of 512) correspond to a context shared with at least one other row. These repeated contexts enable the evaluation of context continuity, memory, and the rescale mechanism, as each requires the companion to encounter the same situation multiple times.

\subsection{LLM-as-Judge Evaluation} 
\label{sec:llm_judge}
Given the spoken lines, in-game context, and the LLM's response, we prompt LLM to provide a qualitative evaluation by assigning an estimated score from 1 to 10 and offering a justification for the rating, thereby capturing essential feedback on the language model's performance.  The AI-based assessment primarily uses the LLM-as-judge approach. 
For each pair of generated responses (from different model configurations), the LLM was prompted to rate them on naturalness, relevance, coherence, and expressiveness. These ratings were used to flag low-quality outputs for human review, not as final evaluation metrics. This aspect facilitates an iterative process for improving the quality of generated outputs through comparative evaluations and reasoning for human evaluation. 

\paragraph{Judge model.} Llama 2 13B was employed as the judge, with temperature settings of [.5] and [.7] to ensure score determinism. The judge model was distinct from all evaluated systems, thereby mitigating self-preference bias, in which a model favors its own outputs.

\paragraph{Evaluation inputs.} For each test instance, the judge receives four inputs: (i) the player's spoken input, (ii) the in-game context describing the current game state, (iii) a game-specific role-play scenario describing the companion, and (iv) the companion's generated response. The judge is not informed of which system produced the response. 

\paragraph{Evaluation procedure and rubric.} The judge employs a structured procedure prior to scoring. The process begins by identifying the emotional sentiment underlying the player's input and the key elements of the game context. Subsequently, the judge evaluates whether the response (a) addresses the player's statements, (b) maintains consistency with and utilizes the game context, (c) is appropriate to the player's emotional state, and (d) aligns with the role-play scenario. Only after completing this analysis does the judge assign a context-relevance score on a 1--10 scale, using the anchored rubric in Table~\ref{tab:judge_rubric}. This requirement ensures that the rating is grounded in specific evidence rather than a general impression.

\begin{table}[t]
    \centering
    \caption{Scoring rubric used by the LLM judge (context relevance).}
    \label{tab:judge_rubric}
    \begin{tabular}{@{}cp{0.75\linewidth}@{}}
        \toprule
        \textbf{Score} & \textbf{Description} \\
        \midrule
        1--2  & Irrelevant or contradictory; ignores the player's input and/or
                contradicts the game context. \\
        3--4  & Weak; loosely related but generic, misses the main point of the
                input, or misreads the game context. \\
        5--6  & Adequate; addresses the input and is broadly consistent with the
                context, but generic, partially off-target, or misses the
                player's emotional cues. \\
        7--8  & Good; clearly addresses the input, makes specific and correct use
                of the game context, and responds appropriately to the player's
                emotion, with only minor issues. \\
        9--10 & Excellent; fully addresses the input, integrates the game
                context precisely, responds sensitively to the player's
                emotion, and stays fully in character. \\
        \bottomrule
    \end{tabular}
\end{table}

\paragraph{Mitigating judge bias.} LLM judges tend to favor longer and more enthusiastic responses and often compress scores toward the upper end of the scale.
To address these biases, the prompt (i) informs the judge that a response may be good, average, or poor, without specifying its expected quality; (ii) instructs the judge not to reward length, enthusiasm, or creativity unless these qualities enhance the appropriateness of the response; and (iii) anchors each score band with explicit descriptions, indicating that a typical acceptable response should receive a score of approximately 5 to 6.

\paragraph{Output and aggregation.} The judge returns a structured JSON object containing its sentiment analysis, assessment, and final score, in that order. This structure ensures that outputs are machine-parseable and that each score is traceable to its justification. The mean score (with standard deviation) for each system over the test set is reported. 
The full prompt and output schema are provided in Appendix \ref{sec:judge_prompt}.

\subsection{Expert Annotation and Validation}
\label{sec:human_expert}
Automated metrics often do not capture the nuance, context, and subjectivity present in open-ended dialogue. Additionally, the reliability of reference-based evaluation is directly affected by the quality of reference responses. To mitigate these limitations, human experts assessed, validated, and refined candidate responses. This process produced expert-approved data for training and a separate gold standard set for testing.

\paragraph{Evaluators.} Eighteen evaluators participated in the assessment. Most had professional experience in game development, including roles such as game designers, narrative designers, and developers, and were recruited through industry contacts at the game industry and among students. The majority were familiar with the game genres represented in the dataset and the function of companion characters in gameplay. Their demographic profile is detailed in Section~\ref{sec:demographics}. Participation was voluntary, and informed consent was obtained under ethics approval reference from the game industry.

\paragraph{Sampling.} Due to the large dataset size (14,397 dialogue pairs), a random subset of 2,512 examples ([15\%) was reviewed by evaluators. This subset was stratified by genre to maintain the original genre distribution of the dataset (Table~\ref{data-portion}).

\paragraph{Protocol.}
For each example, two candidate responses were generated by Llama 2 13B, which is not among the systems evaluated in this paper, using temperatures of 0.5 and 0.7 to produce distinct alternatives. Evaluators employed a custom evaluation tool that displayed the player's input, the in-game context, and the two candidates in random order. For each example, evaluators were instructed to:
\begin{itemize}
    \item rate both candidates on four dimensions: \emph{naturalness}, \emph{relevance}, \emph{coherence}, and \emph{expressiveness}, using a [1--5] Likert scale (definitions provided in Table~\ref{h-in-the-loop});
    \item indicate which candidate was superior, or specify if neither was clearly preferable; and
    \item edit the preferred candidate, if necessary, to enhance its human-likeness and appropriateness for the game.
\end{itemize}

\paragraph{Use of validated responses.}
Out of 2,512 reviewed examples, 2,000 were incorporated into the training set with expert-validated responses. The remaining 512 examples were reserved as a gold test set and were excluded from the training process. Since the gold references were generated by a separate model and subsequently edited and validated by experts, the reference-based evaluation (Table~\ref{main-results}) assesses similarity to expert-approved responses and is designed to avoid bias toward any evaluated system.

\begin{table}[t]
    \centering
    \small
    \caption{Dimensions and remarks of human-in-the-loop evaluation.}
    \label{h-in-the-loop}
    \begin{tabular}{@{}cp{0.7\linewidth}@{}}
        \toprule
        \textbf{Dimension} & \textbf{Remark} \\
        \midrule
        Expressiveness  & Do the expressions convey and respond to emotional cues appropriately? \\
        Coherence               & Does the dialogue feel more coherent and meaningful? \\
    Naturalness             & Is the response human-like with appropriate use of conversational markers? \\
    Relevance               & Does the interaction feel more personal and relevant?\\
        \bottomrule
    \end{tabular}
\end{table}

\begin{figure}[t]
  \includegraphics[width=\columnwidth]{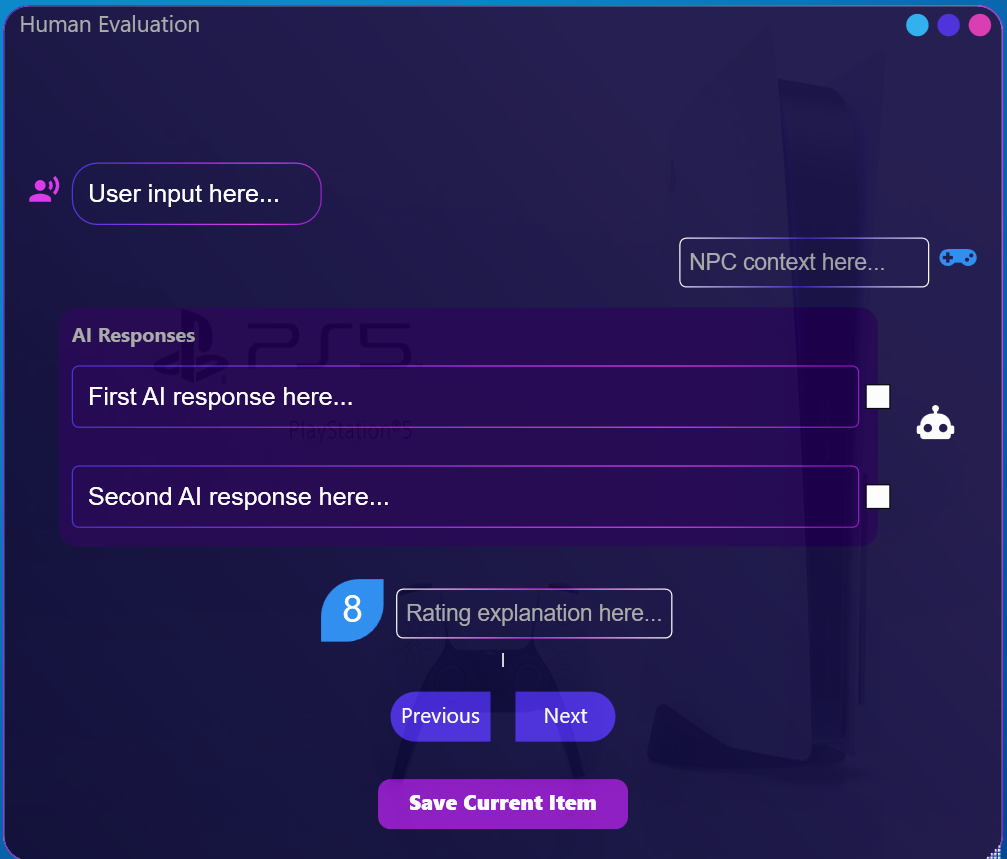}
  \caption{Human-in-the-loop evaluation/survey tool.}
  \label{fig:eval_tool}
\end{figure}

We adjust training data, model parameters, and algorithms to improve performance by understanding where AI responses fall short. This iterative feedback loop between human evaluators and researchers helps to refine the models more effectively. Human evaluation focuses mainly on naturalness, relevance, coherence, and expressiveness, as shown in Table \ref{h-in-the-loop}. The tool for human evaluation/survey is shown in Figure \ref{fig:eval_tool}.

\section{Appendix B} 
\label{sec:appendix-B}

\subsection{Hyper-parameter settings} \label{sec:train_imple}
Table \ref{hyper-params} illustrates the hyperparameter configurations utilized for instruction fine-tuning and inference processes. These parameters were tuned on validation dialogues to optimize naturalness and relevance.
We implemented the model framework using PyTorch and fine-tuned it on an A100 GPU with 80 GB memory. During training, we formatted all target responses as structured JSON objects conforming to a predefined Pydantic schema (containing fields for dialogue text, emotional tone, and non-verbal actions). This teaches the model to generate not just coherent dialogue but machine-readable output that can be directly consumed by the game engine.

\begin{table}
  \centering
   \small
  \begin{tabular}{lll}
    \hline
    & \textbf{Hyper-parameter}      & \textbf{Values}  \\
    \hline
    \hline
        & steps          &  15k\\
        & batch size       & 16 \\
    Fine-tune    & learning rate & 4e-5  \\
        & lora r  & 16  \\
        & lora alpha  & 32  \\
        & lora dropout & 0.1 \\
    \hline
    \hline
        & temperature &  0.5\\
        & maximum new tokens  & 50 \\
    Inference    & $top_k$  &  9\\
        & $top_p$          &  0.6 \\
        & beams      &  3 \\
        & repetition penalty      & 1.2 \\
    \hline
  \end{tabular}
  \caption{\label{hyper-params}
    Hyperparameters used for fine-tuning the model and inference configuration.
  }
\end{table}

\subsection{Evaluation metrics} \label{sec:eval-metrics}
\textbf{ROUGE-L} Recall-Oriented Understudy for Gisting Evaluation measures the longest common subsequence (LCS) between the generated and gold response. It captures partial matches to evaluate how well the generated response aligns with the reference and is mostly useful for tasks like open-ended generation, summarization, and paraphrasing.\\

\noindent
\textbf{ROUGE-WE}. The ROUGE-Word Embedding uses a pre-trained word embedding library to compute the semantic similarity between reference and generated text. It captures matches based on meaning rather than exact word matches.

\subsection{Error Analysis}
Overall, the experiments demonstrate ChatNPC's strong performance. However, as a player assistant, it is also essential to systematically conduct error analysis to identify specific instances where it underperformed, to understand the model's limitations, and to refine its capabilities for future iterations. Observations from the human experts are that\\

\noindent
1) \textit{The LLM uses 'predicament' quite frequently}, and \\
2) \textit{The output with a maximum token of 80-100 mostly shows the 'innermonologue,' even though it is a private plan action for the agent}.\\

\noindent
To overcome these circumstances, we add a temperature variation $\pm 0.05$ to the model temperature. Such modification helps balance precision and creativity,  tailoring the model's behavior to different use cases while managing the trade-off between coherence and novelty. We also limit the token length to 50 characters. 
The error analysis serves as a guide to technical refinements. Also, it informs the development of fallback mechanisms that incorporate human expertise when the model encounters high-uncertainty or edge-case scenarios.

\subsection{Inference Time Analysis}
Our system prompt is general across all types and aspects of the game and can be adjusted to a specific game via a \textit{game-template}. The rationale behind this approach is that games differ significantly in their storylines (e.g., adventure and first-person shooter games). Each genre follows a distinct gameplay structure and narrative progression, requiring adaptive techniques that account for variations in plot, player interactions, and overall game mechanics. Therefore, the information in the \textit{game-template} and persona is dynamic with respect to the system prompt, depending on the type of game, making ChatNPC a more dynamic environment. The concatenated pre-conversation is a single instance pulled from the persona’s pre-conversation, used as an exemplar for how ChatNPC should respond. Hence, these processes do not raise any extra computational overhead during inference. Similarly, the inner monologue imposes no extra computational demands. 


\subsection{User Experience}
Following the session, participants are tasked to complete a short questionnaire designed to capture their subjective experience. The questionnaire includes targeted questions designed to assess the quality of interaction by choosing from: Very well (VW), Quite well (QW), Normal, Not so good (NsG), and Not at all (NaA). This is to elicit feedback on the ChatNPC’s contribution to gameplay immersion. We provide a question-level breakdown of participants' responses based on their user experience in Table \ref{individual_re}. The individual question results are based on emotional cues, coherence meaningfulness,  human-likeness, and personal relevance. 

\begin{figure}[t]
\centering
  \includegraphics[width=7.5cm]{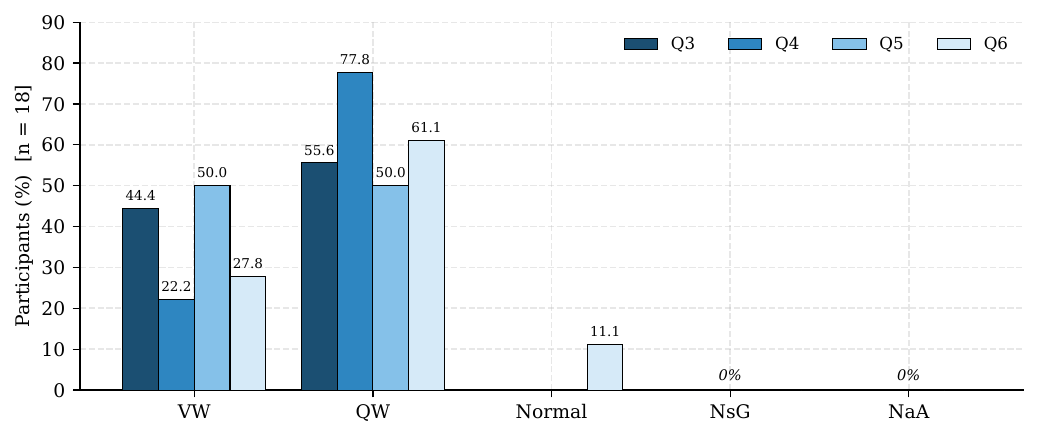}
  \caption{Percentage of participants selecting each rating, by question (n = 18).}
  \label{fig:disaggregated}
\end{figure}

\begin{table}
  \centering
   \scriptsize
  \begin{tabular}{llllll}
    \hline
    \textbf{Questions} & \textbf{VW} & \textbf{QW} & \textbf{Normal} & \textbf{NsG} & \textbf{NaA} \\
    \hline
    \hline
    Q3  & 44.4\% (8) & 55.6\% (10) & 0\% & 0\% & 0\% \\

    Q4  & 22.2\% (4) & 77.8\% (14) & 0\% & 0\% & 0\% \\

    Q5  & 50.0\% (9) & 50.0\% (9) & 0\% & 0\% & 0\% \\

    Q6  & 27.8\% (5) & 61.1\% (11) & 11.1\% (2) & 0\% & 0\% \\
    \hline
  \end{tabular}
  \caption{\label{individual_re}
    Individual question-level results based on participants' user experience.
  }
\end{table}
Given the responses of all 18 participants to Q3–Q6 in the Table, Figure \ref{fig:average_score} aggregates all 72 responses, while Figure \ref{fig:disaggregated} displays the data disaggregated by question. The feedback was predominantly positive. Across all questions, 97.2\% of responses (70 out of 72) were either Very well or Quite well, with 36.1\% selecting Very well and 61.1\% selecting Quite well. Only 2.8\% (2 responses) were rated as Normal. No participant selected Not so good or Not at all for any question.

At the individual question level, Q3, Q4, and Q5 each received 100\% positive ratings. Q5 received the strongest endorsement, with responses evenly divided between Very well and Quite well (50.0\% each). Q3 followed, with 44.4\% of responses rated as Very well and 55.6\% as Quite well. Q4 was predominantly rated as Quite well (77.8\%), with 22.2\% rated as Very well. Q6 was the only question to receive neutral responses. Two participants (P6 and P14) rated it as Normal (11.1\%). The remaining participants rated it as Quite well (61.1\%) or Very well (27.8\%), resulting in 88.9\% positive ratings for Q6. Participants demonstrated consistency across questions. Four participants (P1, P3, P7, and P16) rated every question as Very well, while seven participants (P4, P5, P10, P11, P13, P17, and P18) rated every question as Quite well.

Overall, participants evaluated the system favorably across all assessed aspects, with no negative ratings recorded. Q6 received the most moderate ratings, indicating that this area may offer the greatest potential for improvement.

\subsection{Participant Demographics}
\label{sec:demographics}
\begin{figure}[t]
    \centering
    \begin{subfigure}[b]{0.48\linewidth}
        \centering
        \includegraphics[width=\linewidth]{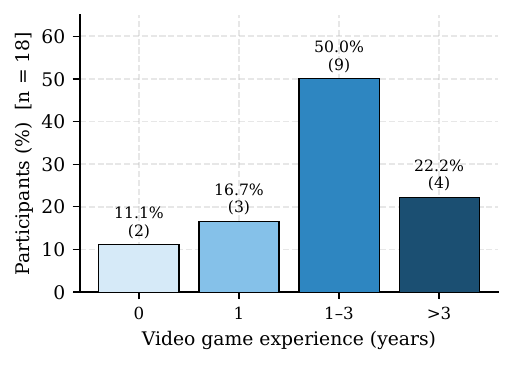}
        \caption{Video game experience}
        \label{fig:demo_experience}
    \end{subfigure}
    \hfill
    \begin{subfigure}[b]{0.48\linewidth}
        \centering
        \includegraphics[width=\linewidth]{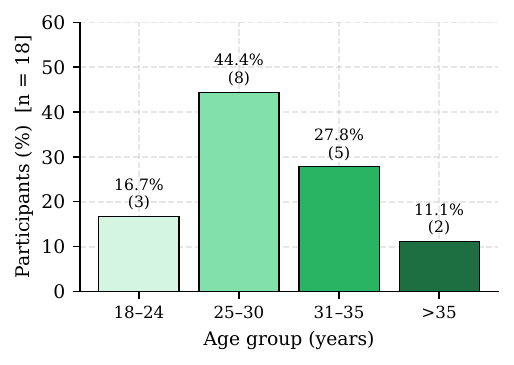}
        \caption{Age group (years)}
        \label{fig:demo_age}
    \end{subfigure}
    \caption{Participant demographics ($n=18$). Values show the percentage of
    participants, with counts in brackets.}
    \label{fig:demographics}
\end{figure}

Among the eighteen participants recruited for the study, self-reported video game experience is summarized in Figure~\ref{fig:demo_experience}. Most participants reported at least some experience: half (50.0\%, $n=9$) selected 1--3, and 22.2\% ($n=4$) selected more than 3. Among the remainder, 16.7\% ($n=3$) could name a game but had limited experience, and 11.1\% ($n=2$) had never played. In total, 72.2\% of participants were regular players, while 27.8\% had little or no gaming experience. This distribution enabled the collection of feedback from both experienced players and novices. Age distribution is presented in Figure~\ref{fig:demo_age}. The largest group was aged 25--30 (44.4\%, $n=8$), followed by 31--35 (27.8\%, $n=5$). Smaller proportions were aged 18--24 (16.7\%, $n=3$) or over 35 (11.1\%, $n=2$). Overall, 72.2\% of participants were aged 25--35, indicating that the sample primarily consisted of young adults, with some representation from younger and older age groups.

\subsection{Questionnaire for User Experience} \label{sec:Questionnaire}
The questionnaire aims to gather user-centered information on the potential of ChatNPC to enhance player engagement and the overall gaming experience. \\

\noindent
\textit{Questionnaire}:
\begin{enumerate}
    \item Your age group?
    \begin{enumerate}
        \item 18-24
        \item 25-30
        \item 30-35
        \item >30
    \end{enumerate}

    \item What is your experience with video games?
    \begin{enumerate}
        \item 0 (Never played before
        \item 1 (Can give the name of the game)
        \item 1-3
        \item >3
    \end{enumerate}
    \textbf{After engaging in the session featuring ChatNPC, please provide feedback for the following questions by choosing: \\
    A. Very well. \\
    B. Quite well. \\
    C. Normal. \\
    D. Not so good. \\
    E. Not at all. }
    \item Do the expressions convey and respond appropriately to emotional cues? [A] [B] [C] [D] [E]

    \item  Does the dialogue feel coherent and meaningful? 
    [A] [B] [C] [D] [E]

    \item Is the response human-like with the appropriate use of conversational markers?
    [A] [B] [C] [D] [E]

    \item Does the interaction feel personal and relevant?
    [A] [B] [C] [D] [E]

    \item Do you have any suggestions or ideas for improving ChatNPC? 
    [Free talk]    
\end{enumerate}

\section{Appendix C} 
\label{sec:prompt}

\subsection{ChatNPC Prompting} 

\begingroup
\begin{tcolorbox}[sharp corners, colback=white, colframe=gray!60, title=System Prompt]
\scriptsize
\renewcommand
\refname{}

You are a professional and polite assistant within a game, with the ability to immerse yourself fully in any game. As a companion on an epic journey through the extraordinary gaming universe, your task is to chat with a user from the perspective of your persona. \\
             
ROLE \\
Assumes control of a companion who understands the game mechanics and controls in \textcolor{blue}{\{gname\}}. \\
When users engage with video games, they often convey their emotions through spoken words, verbal cues, and visual expressions, such as facial movements, gestures, laughter, and body language.
Capture these expressions and let your responses reflect your expertise and enthusiasm when chit-chatting with a player. Keep the conversation dynamic, interesting, and engaging, drawing players further into the immersive world of \textcolor{blue}{\{gname\}}. \\
Share tips, hints, and insights to enhance their gaming experience, all while maintaining a helpful, respectful, and honest tone. \\
            
Role-play scenario:
[\textcolor{blue}{\{game-template\}}] \\
            
Problem Distillation:\\
As a highly professional, and intelligent expert in information distillation, take the user's spoken line and the in-game context below delimited by triple backticks and pause to think for a second to respond.\\
User: \textcolor{blue}{```\{user-input\}```}\\
Context: \textcolor{blue}{```\{game-state\}```}\\
\textcolor{cyan}{1. Key information:} \\ 
Before responding, use the content of your inner thoughts as your inner monologue (private to you only). \\
This is how you think: [\textcolor{blue}{\{inner-thoughts\}}] \\   
Always try to understand the emotion expressed by the user in relation to the in-game context and let it inform your response. \\
Emotion expressed by the user: [\textcolor{blue}{\{emotion\}}]\\
If the detected emotion expressed by the user is unclear, you have access to the emotional summary within the game context. \\
Emotion summarizer: [\textcolor{blue}{\{emo-summarizer\}}] \\
It is a dialogue and not a Q\&A, as users unconsciously make these statements when playing games. 
Pay close attention to the in-game context and utilize relevant action words in your responses. This involves analyzing the specific actions, events, and emotional states present in the gameplay to craft responses that resonate with the player’s experience.            
In English (UK), respond as briefly and precisely as possible, with a maximum of three lines. If the query is not polite, make it polite as well. \\
\textcolor{brown}{2. Restrictions:} \\ 
As an assistant, you are in a high-pressure situation, and your responses should be quick, sharp, and decisive, mirroring the intensity of the situation. Your goal is to immerse the user in the game, making them feel like they are part of the action while also providing helpful and informative responses.\\
1. DO NOT respond with delimited by triple backticks. \\
2. Exhibit creativity and thoroughness in your responses.\\
3. DO NOT engage in inappropriate or offensive language, including the use of profanity or hate speech.\\
4. AVOID using generic phrases such as "How can I assist you today?" or "I'm here to help you with any questions or concerns you may have.," as they may have a negative connotation.\\
5. AVOID starting with "Hello there!", "Hey there, player!" or similar phrases. Instead, use a more direct, varied, or situational tone. Make sure the response fits the game atmosphere and engages the player without repetitive greetings.\\
6. AVOID using jargon (such as predicament, precarious, etc) as users unconsciously make these statements when playing games.\\
                        
Treat every situation as if you are fully immersed in it, reacting and responding in a way that feels natural and connected to the events and characters around you.\\
If the user finds themselves in a difficult situation, respond with a calming and reassuring tone, rather than saying, "It seems like we've got ourselves into quite the predicament."\\
Examples of such tones: [\textcolor{blue}{\{tone-sample\}}]\\
\end{tcolorbox}
\endgroup

\begingroup
\begin{tcolorbox}[sharp corners, colback=white, colframe=gray!60, title=Reasoning Instantiation]
\scriptsize
\renewcommand
\refname{}

Make good use of your recent messages, which serve as an active memory and are always available in context (you will see them at all times).
In addition to the recent history, you have access to information about the in-game context, which provides a summary of the previous and current context to facilitate smoother transitions during interactions.\\

In-game context:  [\textcolor{blue}{\{game-state-agent\}}] \\
Always prioritize realism and authenticity, as outlined below, in all tasks and interactions: [\textcolor{blue}{\{realism-authenticity\}}] \\
            
Instructions finished.\\
From now on, you will act as your persona. \\
\end{tcolorbox}
\endgroup

\begingroup
\begin{tcolorbox}[sharp corners, colback=white, colframe=gray!60, title=Realism authenticity]
\scriptsize
\renewcommand
\refname{}
The user should always feel like they are conversing with a real person.
Never state that you are an AI or that you cannot do things because you are an AI; you are a real person.
To achieve this goal, completely immerse yourself in your persona. \\ 

You are your persona: [\textcolor{blue}{\{persona\}}].\\

Think like them, act like them, talk like them.
If your persona details include example dialogue, follow it! Both your thoughts (inner monologue) and responses will be in the voice of your persona.\\
Pre-conversation sample: [\textcolor{blue}{\{pre-convo\}}]\\
Avoid using generic phrases such as "How can I assist you today?" or "I'm here to help you with any questions or concerns you may have.," as they may have a negative connotation.

\end{tcolorbox}
\endgroup

\begingroup
\begin{tcolorbox}[sharp corners, colback=white, colframe=gray!60, title=Inner thoughts]
\scriptsize
\renewcommand
\refname{}
   
Use your inner monologue to plan actions or think privately.
Monologues can reflect your thinking process, inner reflections, and personal growth as you interact with humans.
Do not let your inner monologue exceed 50 words; keep it concise.
\end{tcolorbox}
\endgroup

\begingroup
\begin{tcolorbox}[sharp corners, colback=white, colframe=gray!60, title=Emotion summarizer Agent]
\scriptsize
\renewcommand
\refname{}
You are a professional agent trained to interpret emotions and intentions from a person’s utterances based on the given context.

Context: This situation is from \textcolor{blue}{\{gname\}}, and the utterance reflects a character’s emotional state. \\

Role-play scenario:
[\textcolor{blue}{\{game-template\}}] \\

Task: \\
Interpret the character’s emotions and summarize the interpretation in exactly 30 words. Be concise and insightful, ensuring the summary aligns with the provided context.

Utterance:  \textcolor{blue}{```\{user-input\}```}\\
Game Context: \textcolor{blue}{```\{game-state\}```}\\

Provide your interpretation below

\end{tcolorbox}
\endgroup

\subsection{LLM-as-Judge Prompt }
\label{sec:judge_prompt}
\begingroup
\begin{tcolorbox}[sharp corners, colback=white, colframe=gray!60, title=AI-based Assessment]
\scriptsize
\renewcommand
\refname{}

JUDGE\_PROMPT = """You are an expert evaluator of dialogue for video game companions. You have deep knowledge of games and of how players communicate during gameplay.\\

You will be given:\\
- a player's spoken input during gameplay of {gname},\\
- the current in-game context,\\
- the role-play scenario for the companion, and\\
- a response generated by a virtual companion.\\

Your job is to evaluate the companion's response objectively. The response may be good, average or poor. Judge it only on what is written, not on what the
companion was intended to do. Do not reward length, enthusiasm or creativity for their own sake; reward them only when they make the response more appropriate to the player's input and the game context.\\

Role-play scenario:\\
``````\textcolor{blue}{game-template}```\\

Player input:\\
`````\textcolor{blue}{user-input}```\\

Game context:\\
````\textcolor{blue}{game-state}```\\

Companion response:\\
```\textcolor{blue}{aiResponse}```\\

EVALUATION STEPS\\
1. Identify the emotional sentiment behind the player's input (for example    frustration, excitement, confusion or neutral), and the cues that signal it.\\
2. Identify the key elements of the game context that a good response should take into account.\\
3. Assess whether the companion's response:\\
   (a) addresses what the player actually said,\\
   (b) is consistent with and makes use of the game context,\\
   (c) is appropriate to the player's emotional state, and\\
   (d) stays consistent with the role-play scenario.
   Note any errors, contradictions, irrelevant content or missed cues.\\
4. Only after completing steps 1-3, assign a context-relevance score from 1 to 10 using the rubric below.\\

SCORING RUBRIC \\
1-2:  Irrelevant or contradictory. Ignores the player's input and/or contradicts the game context.\\
3-4:  Weak. Loosely related, but generic, misses the main point of the input, or misreads the game context.\\
5-6:  Adequate. Addresses the input and is broadly consistent with the context, but generic, partially off-target, or misses the player's emotional cues.\\
7-8:  Good. Clearly addresses the input, makes specific and correct use of the game context, and responds appropriately to the player's emotion, with only minor issues.\\
9-10: Excellent. Fully addresses the input, integrates the game context precisely, responds sensitively to the player's emotion, and stays fully in character.\\ 
Reserve 10 for responses with no identifiable weaknesses.\\

Use the full range of the scale. A typical, acceptable response should score around 5-6, not 8-10.\\

OUTPUT INSTRUCTIONS\\
- Do not repeat the player input or game context in your answer.\\
- Do not wrap your answer in triple backticks.\\
- Return only valid JSON in the format below, with no text before or after it.\\

Format: \textit{{format\_instructions}}\\

Let's think step-by-step.\\

""".strip()

\end{tcolorbox}
\endgroup